\documentclass{article}
\usepackage{ijcai20}

\usepackage{times}
\usepackage{soul}
\usepackage{url}
\usepackage[hidelinks]{hyperref}
\usepackage[utf8]{inputenc}
\usepackage[small]{caption}
\usepackage{graphicx}
\usepackage{amsmath}
\usepackage{amsthm}
\usepackage{booktabs}
\usepackage{algorithm}
\usepackage{algorithmic}
\usepackage{theorem}

\usepackage{amsfonts}
\usepackage{subfigure}
\title{IC neuron: An efficient unit to construct neural networks}

\author{
	Junyi An$^1$\and
	Fengshan Liu$^1$\and
	Jian zhao$^2$\and
	Furao shen$^1$\footnote{Contact Author}\\
	\affiliations
	$^1$Department of Computer Science and Technology, Nanjing University, Nanjing, China\\
	$^2$School of Electronic Science and Engineering, Nanjing University, Nanjing, China\\
	\emails
	\{junyian, liufengshan\}@smail.nju.edu.cn,\\
	\{frshen, jianzhao\}@nju.edu.cn
}

\begin{document}
	
	\maketitle
	
	\begin{abstract}

		As a popular machine learning method, neural networks can be used to solve many complex tasks. Their strong generalization ability comes from the representation ability of the basic neuron model. The most popular neuron is the MP neuron, which uses a linear transformation and a non-linear activation function to process the input successively. Inspired by the elastic collision model in physics, we propose a new neuron model that can represent more complex distributions. We term it Inter-layer collision (IC) neuron. The IC neuron divides the input space into multiple subspaces used to represent different linear transformations. This operation enhanced non-linear representation ability and emphasizes some useful input features for the given task. 

    We build the IC networks by integrating the IC neurons into the fully-connected (FC), convolutional, and recurrent structures. The IC networks outperform the traditional networks in a wide range of experiments. We believe that the IC neuron can be a basic unit to build network structures.

	\end{abstract}
	
	\section{Introduction}

\begin{figure*}[]
	\centering
		\subfigure[]{
			\centering
			\includegraphics[scale=.20]{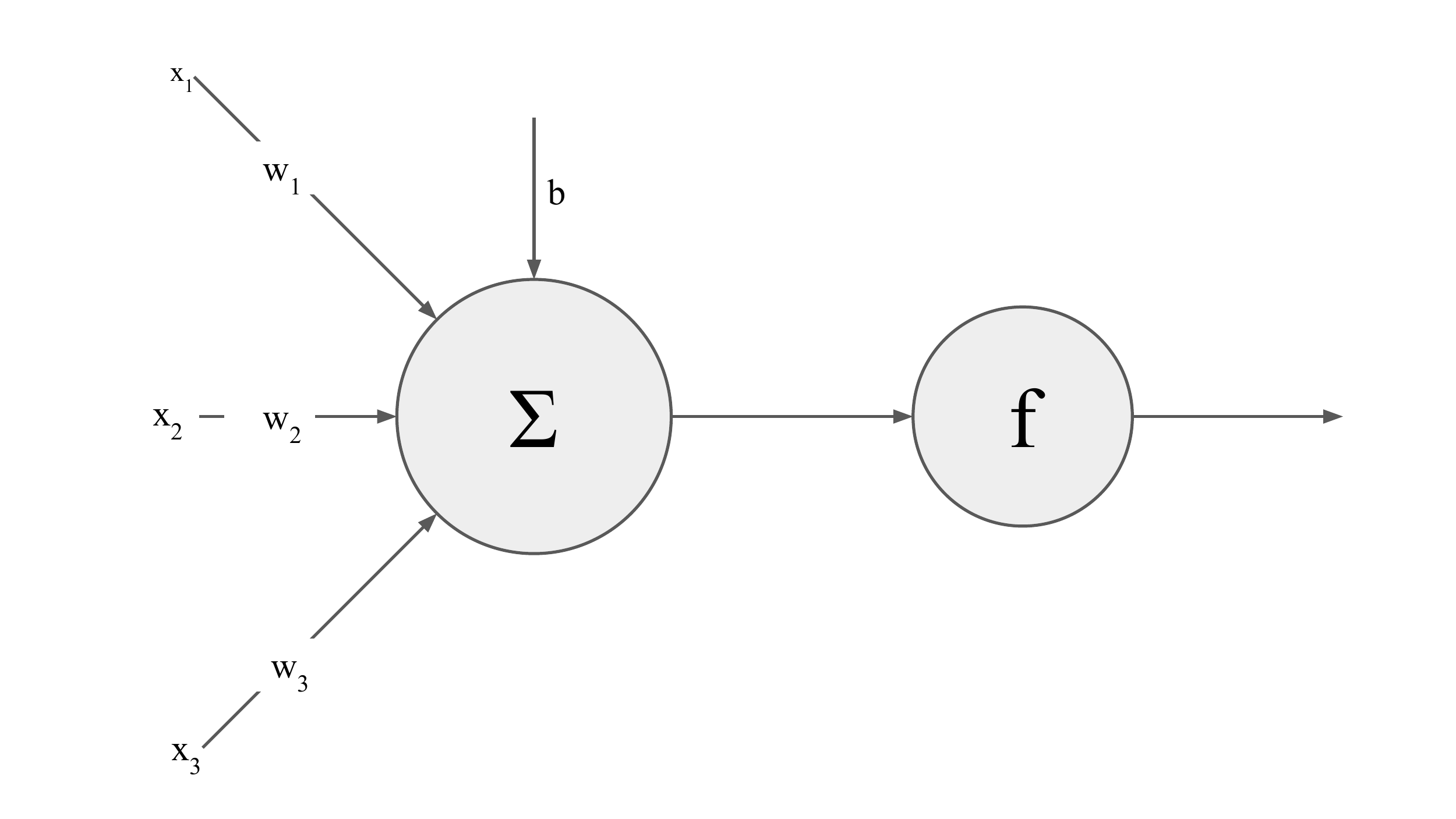}
		}
		\subfigure[]{
			\centering
			\includegraphics[scale=.20]{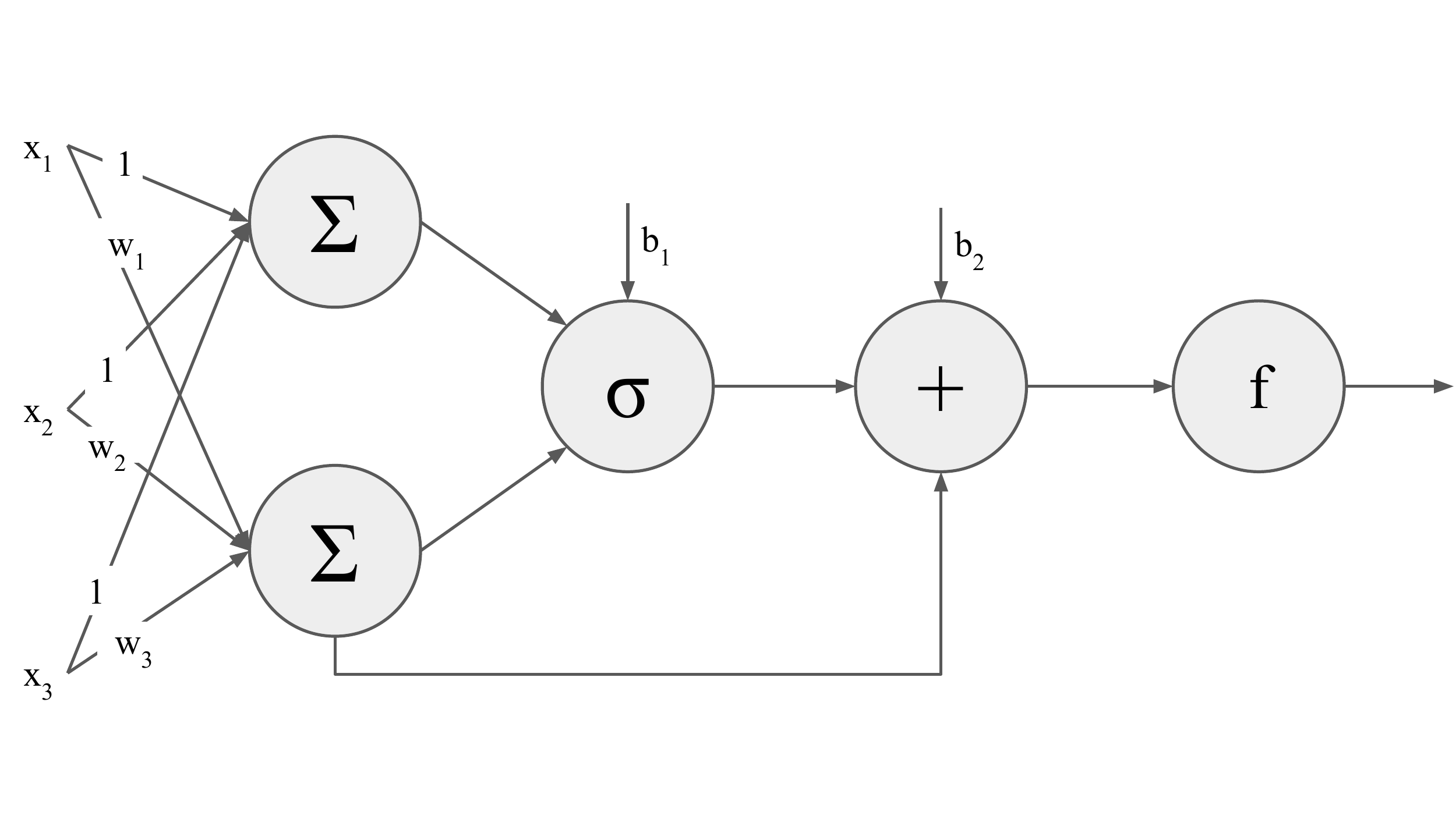}
		}
		\subfigure[]{
			\centering
			\includegraphics[scale=.20]{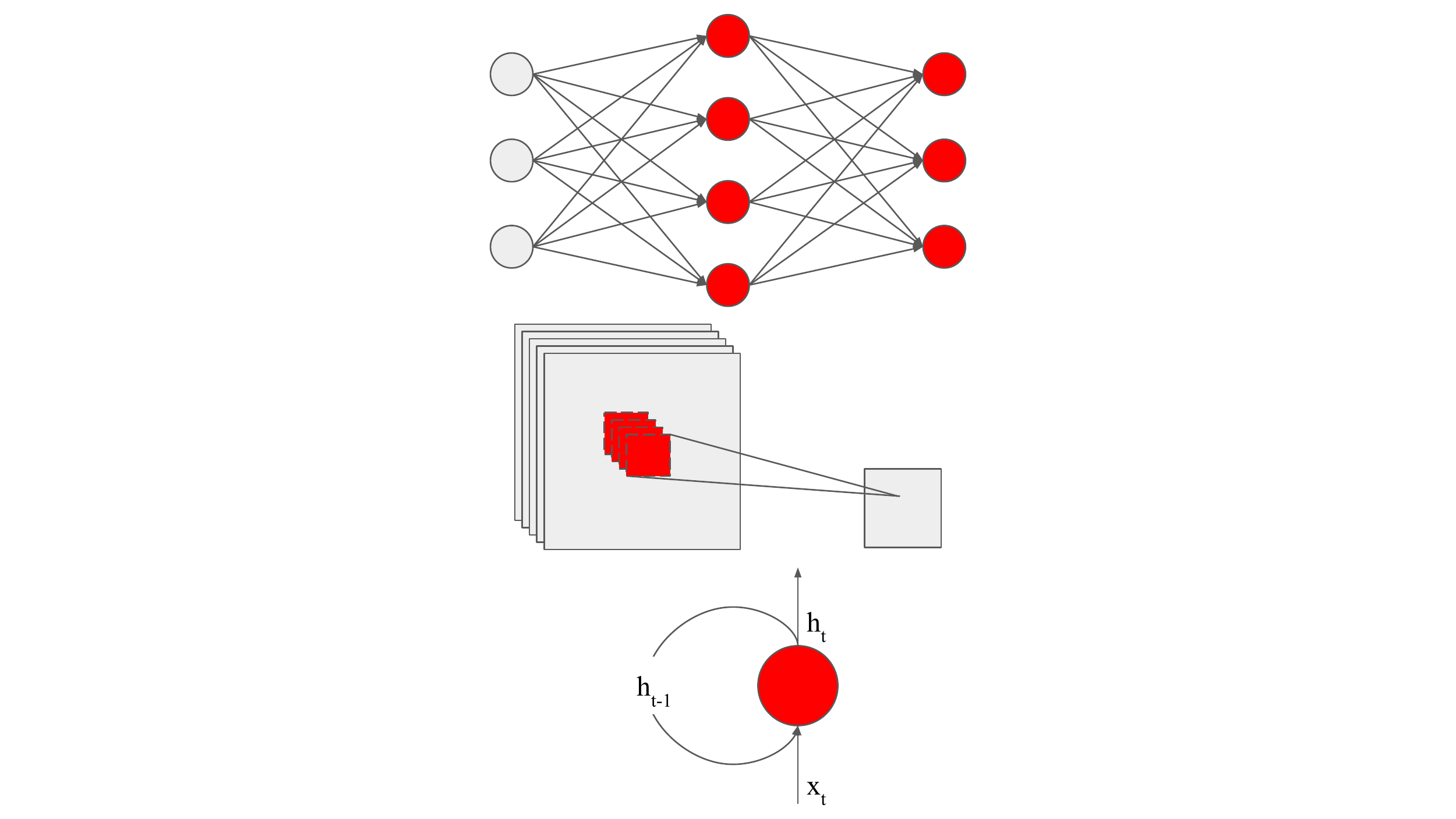}
		}
	
\caption{(a) The MP neuron, $x_{i}$ denotes the input signal, $w_{i}$ denotes connection weight and $b$ denotes bias; (b) The IC neuron, $\sigma$ denotes a non-linear operation, $w'$ denotes an adjustment weight and $+$ denotes addition; (c) Three network structures with the IC neurons. The red regions represent the computational structure of the IC neurons.}
	\label{fig1}
\end{figure*}

	Artificial neural netowrk (ANN) is an important branch of machine learning. The MP neuron~\cite{mcculloch1990logical} as the the basic unit of ANN is the most commonly used neuron stucture. As shown in Figure \ref{fig1}(a), the MP neuron can be formulated as $ y = \mathrm{f}(\sum_{i=1}^{n} \ w_{i}x_{i} + b) $, where a linear transformation and a non-linear activation function are applied to the input successively. This structure enables the netowrk to represent the non-linear distribution of the input signal. Currently, most researchers are concentrating on changing the connections between MP neurons to develop new network structures, of which fully-connected (FC) neural network, convolutional neural network (CNN)~\cite{krizhevsky2017imagenet} and recurrent neural network (RNN)~\cite{graves2013speech} are the most popular ones. However, the developing of new neuron stuctures has not draw much attention. 

The MP neurons are proposed by simply imitating the bihavior of biological neurons. Although some work try to introduce more complex biological mechanisms \cite{maass1997networks} \cite{querlioz2013immunity}, the biological neurons are so complicated that we belived the current arificial neuron is far from reaching the full potental of a biological neuron \cite{ostojic2011spiking}. Therefore, exploring other artificial neuron sturctues become an important topic. 

In this work, we develop a novel artificial neuron model which inspired by the way a biological neuron transmit information to another neuron. We know that a neuron transmit information to a neighboring neuron through releasing neurotransmitters. The neurotransmitters travel accross the space between the two neurons before caught by the neighboring neuron. We find that this process has many similarites with the elastic collision model in physics. Inspired by this, we use the elastic collision model to simulate the way biological neurons transmit information. Based on this, we developed a new aritficial neuron model, which we name the Inter-layer Collision (IC) neuron.

The IC neuron imitates the mathematical form of the elastic collision model, transforming input signal like the process of changing the state of objects during a collision. In addition to the physical significance, we give our mathematical analysis to show how the IC neuron works. The structure of the IC neuron is depicted in Figure \ref{fig1}(b), where scalar $w'$ is introduced to adjust a new input branch and $\sigma$ denotes a non-linear operation with Rectified Linear Unit (ReLU)~\cite{nair2010rectified}. When an input signal passes through this part, the input space is divided into multiple subspaces which can represent different linear transformations. Moreover, the IC neuron retains the advantages of a linear transformation of MP neurons, mainly in the independence of connection weights. Both the IC neuron and MP neuron can map the input with any dimension to a one-dimensional (1-D) output. We prove that in limited input space, the IC neuron can perfectly represent all the distributions that MP can represent. In this way, the IC neuron can replace MP neuron and learn a more abstract distribution of input signals. 

Since the IC neuron has the same format of input and output as the MP neuron, we can easily integrate it into some existing network structures, including the FC, convolutional, and recurrent structures. As shown in Figure \ref{fig1}(c), the framework of the three networks with the IC neurons is similar to traditional networks. The IC neuron can improve the generalization ability of the networks by optimizing the non-linear representation ability of the basic computing unit without changing the network connections. Besides, unlike some methods that use non-linear kernel functions to improve non-linear representation ability~\cite{wang2019kervolutional}, the networks with the IC neurons do not map inputs to a high-dimensionalspace, making it have a small amount of calculation and parameters, which is close to traditional networks. We design experiments for IC networks with the three connection structures. Our experiments show that, in a variety of classification tasks, the IC networks not only exceed the traditional networks with the same hyperparameters in accuracy but also show the advantage of accelerating the training process. Furthermore, the IC networks can achieve better accuracy with a smaller scale, which shows that the IC neuron is an effective basic building unit for neural networks. 

Our contribution can be summarized as follows: 

\begin{itemize}
	\item Inspired by the physical collision model, we propose the IC neuron as a basic building block for neural networks. We prove that the IC neuron can represent more complex distributions compare to the commonly used MP neuron. 
	\item The IC neuron we proposed is a basic computational unit that can be integrated into the majority of existing network structures. We propose strategies to combine the IC neuron with FC, convolutional and recurrent structures, which are popular network structures.  Our research shows that using the IC neuron and the strategies can build more efficient networks.
\end{itemize}

We will introduce the details of the IC neuron and the networks with IC neurons in section 2. We construct a series of experiments to investigate the effectiveness of the IC neuron in section 3. Finally, we summarize this study in section 4.

\section{The Inter-layer Collision Model}

The IC neuron is inspired by the elastic collision model. In this section, we first describe a physical collision scene that reveals the law of physical quantity changes. We then introduce the IC neuron and its physical significance. Section 2.3 gives a mathematical analysis of why the IC neuron has a stronger representation ability than the MP neuron. Finally, we build the networks with the IC neurons and analyze the time and space complexity of them.

\subsection{The physical elastic collision scene}

Physical models often reveal the update or transfer process of physical quantities, e.g. the elastic collision model can clearly indicate the changes in the momentum and speed of objects. Considering an ideal physical environment with only collision forces. Two objects move in a 1D space where we use the left $(-)$ and the right $(+)$ to indicate the direction. The mass of them are $m_{1}$ and $m_{2}$, respectively. Initially, $m_{1}$ lies to the left of $m_{2}$ and they both have a velocity of zero. When $m_{1}$ is given a positive speed $v_{1}$ toward $m_{2}$, according to the laws of energy conservation and momentum conservation, the velocity of $m_{1}$ and $m_{2}$ after collision are:

\begin{equation}
\label{eq1}
\begin{aligned}
&v_{1}' = \frac{m_{1}-m_{2}}{m_{1}+m_{2}}v_{1}, v_{2}' = \frac{2m_{1}}{m_{1}+m_{2}}v_{1}. \\
\end{aligned} 
\end{equation}

According to the above analysis, the result of this collision process is that the object $m_{2}$ will always move to the right ($+$) while the direction of the object $m_{1}$ depending on the quantitative relationship between $m_{1}$ and $m_{2}$. To effectively observe this result, we can place an observation platform on the far right of the above physical scene to obtain the speed value. Furthermore, we can treat this physical model as an information transmitting system where we only care about the overall information flowing out of the system given the input information. In this case, the input information is $v_{1}$ and the output information is $v'' = v_{1}'' + v_{2}''$. Here $v_{1}''$ and $v_{2}''$ are given by:
\begin{equation}
\label{eq2}
\begin{aligned}
v_{1}'' = \sigma \left((w-1)v_{1} \right),  v_{2}'' = wv_{1}, 
\end{aligned}
\end{equation}

where $w$ denotes the coefficient $\frac{2m_{1}}{m_{1}+m_{2}}$, $\sigma$ denotes the ReLU function which is used to get the right component $(+)$ of $v_{1}'$ (Capture objects that continue to move to the right). We notice that $w$ acts as a parameter controlling how much information could be transmitted. In the way, if we make $w$ learnable like what we do in machine learning and add this structure into the neural network framework, the system can be tuned to transmit useful information to the subsequent layers. Based on this assumption, we build a learnable model. To standardize the notation, we replace the ${v_{1}, v''}$ with ${x, y}$ to denote the input and output, respectively. The mathematical form is given by:
\begin{equation}
\label{eq3}
\begin{aligned}
\centering
y = wx + \sigma\left((w-1)x \right).
\end{aligned}
\end{equation}
Inspired by this propagation model, we build the basic IC computational unit introduced in the next section. 

\subsection{The basic IC computational unit}

The process of velocity physical quantity transmission formulated in eq. \ref{eq3} represents a single input and single output model. When the single unit receives multiple inputs, we define the output by adding the results of each input, which can be viewed as the accumulation of multiple collision results on the same object. We combine this process with the activation operation. Considering a n-D input and a bias term, the mathematical form is given by:

\begin{equation}
\label{eq4}
\begin{aligned}
\centering
y &=  \mathrm{f} \left( \sum_{i=1}^{n} \ w_{i}x_{i} + \sigma \left(\sum_{i=1}^{n} \ (w_{i} - 1)x_{i} + b_{1} \right) + b_{2} \right) \\
&= \mathrm{f} \left( \sum_{i=1}^{n} \ w_{i}x_{i} + \sigma \left(\sum_{i=1}^{n} \ w_{i}x_{i} - x_{sum} + b_{1} \right) + b_{2} \right) ,\\
\end{aligned}
\end{equation}

where $\mathrm{f}$ is used to denote an activation function, such as sigmoid or ReLU function. $b_{1}$ and $b_{2}$ are two independent bias used to adjust the center of the model distribution. $x_{sum}=\sum_{i=1}^{n} \ x_{i}$ represents the sumation of all features of input. Eq. 4 defined the basic version of the IC neuron. In the physical scene constructed in 3.1, we determine whether the object $m_{1}$ passes through the right observation platform by controlling the relationship of mass. This can be regarded as a screening process that observers always hope to retain objects that carry key information instead of being disturbed by irrelevant ones. Corresponding to the IC neuron, the input signal is divided into $\sum_{i=1}^{n} \ w_{ij}x_{i} + b_{2}$ and $\sigma \left(\sum_{i=1}^{n} \ (w_{i} - 1)x_{i} + b_{1} \right)$. The term $\sigma \left(\sum_{i=1}^{n} \ (w_{i} - 1)x_{i} + b_{1} \right)$ can be explained by physical significance of the object $m_{1}$, which enhances the key information and suppress the useless one. In addition to fitting features, IC neurons play a role in controlling the flow of information, capturing the fine-gain features through multilayer framework, which are beneficial for a given task.  
In order to further improve the non-linear representation ability of IC neurons, we relax the constant $1$ in $\sigma \left(\sum_{i=1}^{n} \ (w_{i} - 1)x_{i} + b_{1} \right)$ to a learnable value $w'$. The formulation is given by:

\begin{equation}
\label{eq5}
\begin{aligned}
\centering
y &=  \mathrm{f} \left( \sum_{i=1}^{n} \ w_{i}x_{i} + \sigma \left(\sum_{i=1}^{n} \ (w_{i} - w')x_{i} + b_{1} \right) + b_{2} \right) \\
&= \mathrm{f} \left( \sum_{i=1}^{n} \ w_{i}x_{i} + \sigma \left(\sum_{i=1}^{n} \ w_{i}x_{i} - w'x_{sum} + b_{1} \right) + b_{2} \right) , \\
\end{aligned}
\end{equation}

$w'_{j}$ is named the adjustment weight. It can be regarded as the intrinsic weight of one neuron, which is different from the weight $w_{j}$ connecting two neurons. We term the new structure the standard IC neuron, which not only preserves the physical significance of the elastic collision model but also adds more flexibility in controlling the information transmission. We prove that the introduction of $w'_{j}$ can alleviate the restriction on non-linear representation. 

\begin{figure*}[]
	\centering
	\subfigure[]{
		\includegraphics[height=3.5cm, width=3.5cm]{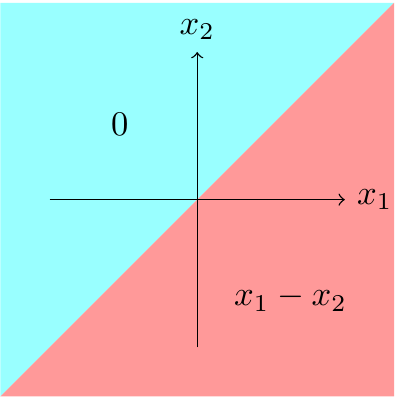}
	}
	 \hspace{1cm}
	\subfigure[]{
		\includegraphics[height=3.5cm, width=3.5cm]{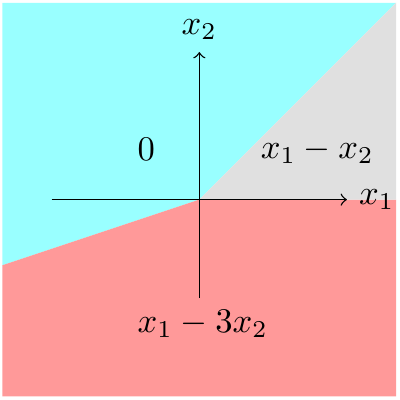}
	}
	 \hspace{1cm}
	\subfigure[]{
		\includegraphics[height=3.5cm, width=3.5cm]{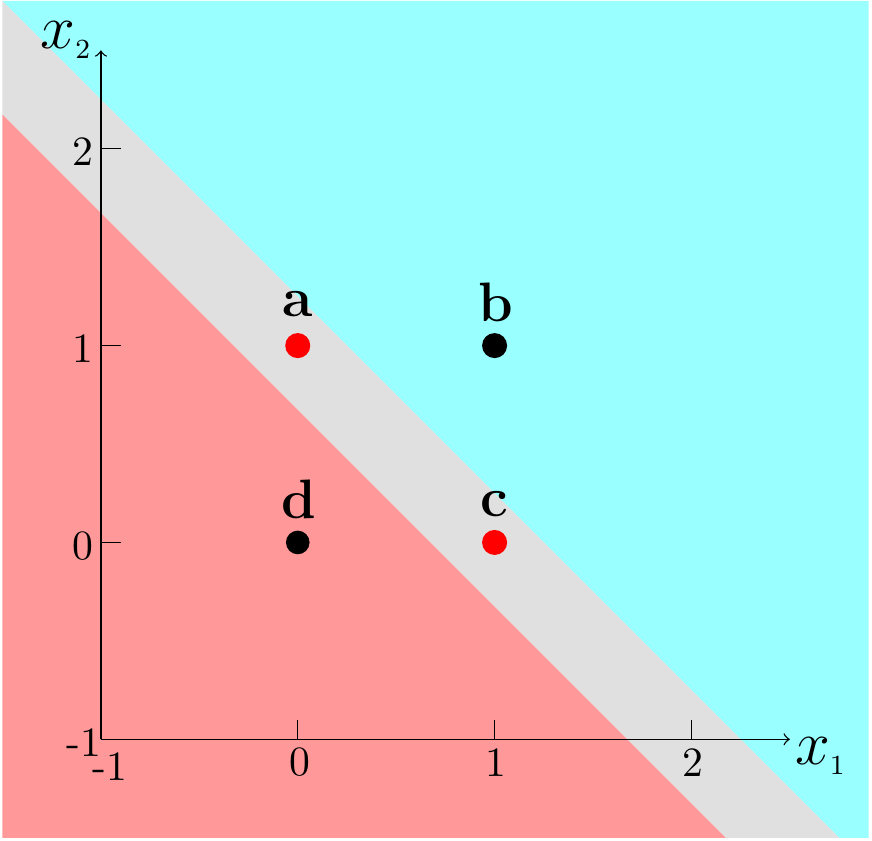}
	}
	\caption{The value of $f(x_{1},x_{2})$ given $x_{1}, x_{2}$. (a): $f(x_{1},x_{2})=\sigma (x_{1}-x_{2})$. (b): $f(x_{1},x_{2})=\sigma (x_{1}-x_{2}+\sigma(-2x_{2}))$. (c): $f(x_{1},x_{2})=\sigma (w_{1}x_{1}+w_{2}x_{2}+b_{1}+\sigma((w_{1}-1)x_{1}+(w_{2}-1)x_{2}+b_{2}))$. Here $w_{1} = w_{2} = 0.2805$. $b_{1} = -0.3506$ and $b_{2} = 0.6463$ are used to shift the boundary across the whole space. }\label{fig2}
\end{figure*}

\subsection{The mathematical analysis of the IC neuron}

\textbf{Multiple linear transformation.} We first consider the representation ability of the basic IC neuron. Different from traditional neurons $y=f(\sum_{i=1}^{n} \ w_{i}x_{i})$, eq. 4 uses a non-linear representation instead of a linear one inside an activation function. We use a term $H = \sum_{i=1}^{n} \ (w_{i} - w')x_{i}$ represents a hyperplane in a $N$-dimensional Euclidean space, which divides Eq. 4 as follows:

\begin{equation}
 \label{eq6}
y =
\begin{cases}
\mathrm{f} \left( 2\sum_{i=1}^{n} \ w_{i}x_{i} - \sum_{i=1}^{n} \ x_{i} \right) & \text{if } H \ge 0\\
\mathrm{f} \left(\sum_{i=1}^{n} \ w_{i}x_{i} \right) & \text{if } H < 0
\end{cases}.
\end{equation}

Intuitively, this IC neuron has a stronger representation ability than the MP neuron, since it can produce two different linear representations instead of one before the activation operation $\mathrm{f}$. To visually distinguish the difference between the two kinds of neurons, we map the 2-D data distribution onto the plane and use ReLU as the activation function to show how the two neurons generate non-linear boundaries. Figure \ref{fig2}(a, b) shows that the single MP neuron and the IC neuron divide the two-dimensional Euclidean space into multiple subspaces, each of which can represent a fixed linear transformation or a zero transformation. We observe that a single IC neuron can divide one more subspace to represent a different linear transformation. Furthermore, we map the XOR problem which is a typical linear inseparable problem into a two-dimensional space to explain the difference. To distinguish between $(1,0),(0,1)$ and $(1,1),(0,0)$, the four points need to be divided into at least three linear subspace. For the ReLU MP neuron, the non-linear boundary is a straight line. At least two neurons are required to solve the XOR problem. However, the non-linear boundary of the IC neuron is a broken line, providing a single neuron possibility to solve the XOR problems. Figure \ref{fig2} gives a solution of single IC neuron, dividing into three-space where $(0,0),(1,1)$ are in the zero space and $(1,0),(0,1)$ is in two spaces with similar representation. 

\begin{figure*}[]
	\centering
	\subfigure[]{
		\centering
		\includegraphics[scale = 0.15, trim={20cm 0 20cm 0},clip]{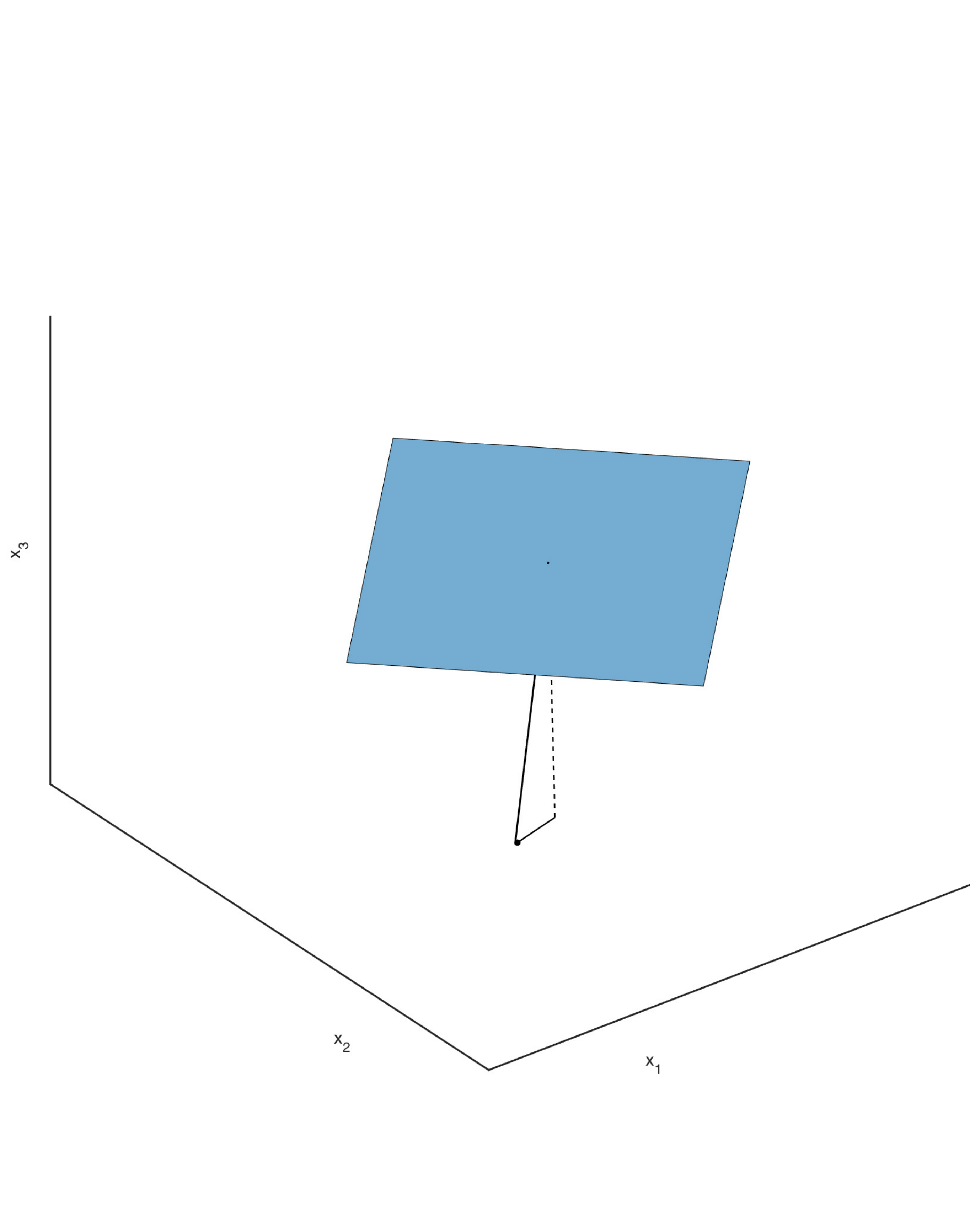}
	}
	\hspace{3cm}
	\subfigure[]{
		\centering
		\includegraphics[scale = 0.15, trim={20cm 0 20cm 0},clip]{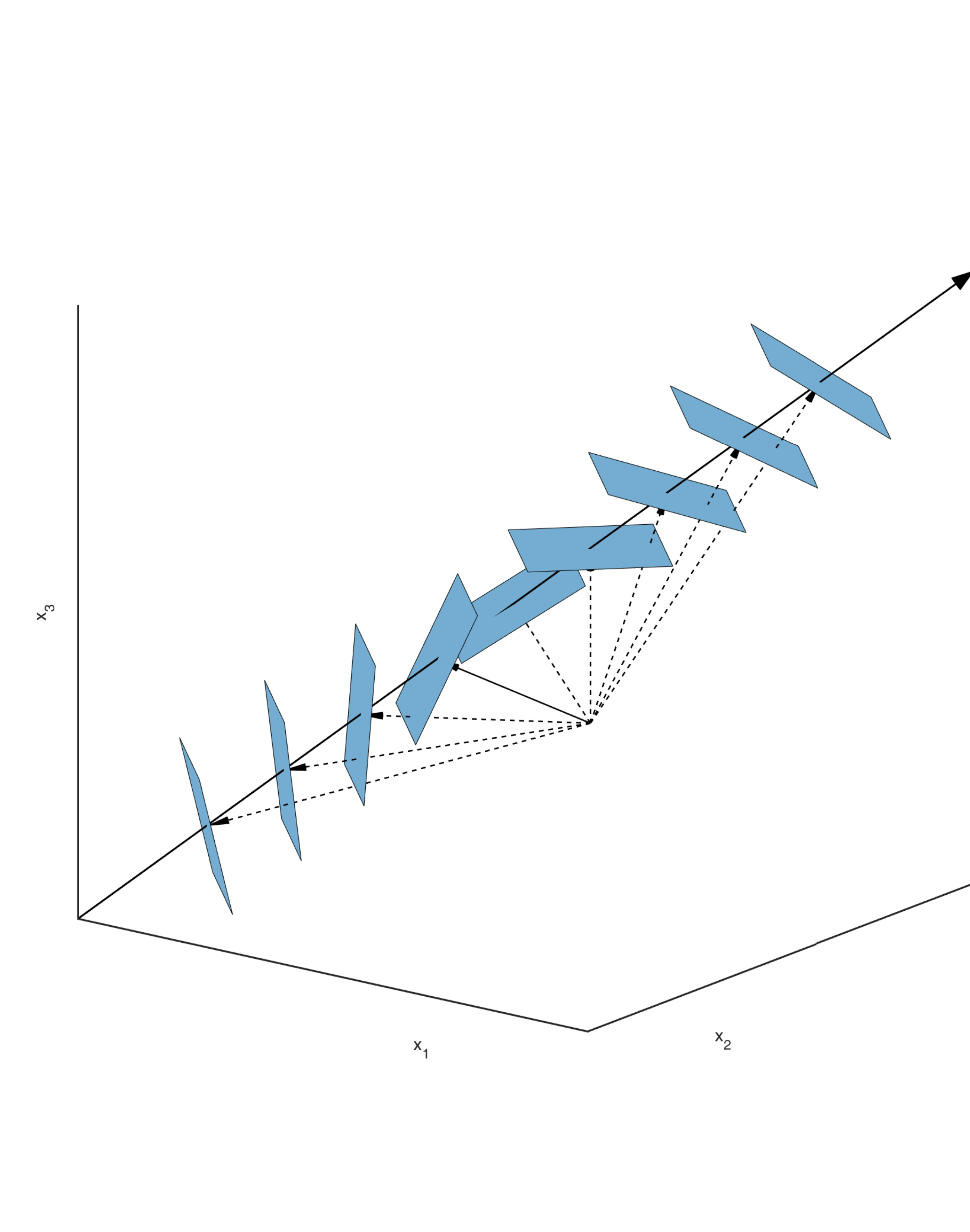}
	}
	\caption{(a): The hyperplane $H=0$ dividing the 3-D input space; (b): The rotation process of the hyperplane according to the adjustment weight $w'$.}
	\label{fig3}
\end{figure*}

\textbf{Adjustment weight.} Although the basic neuron uses the hyperplane $H=0$ to increase the number of patterns, the calculation of the hyperplane is limited by weights $\mathbf{w_{i}}$, making it not flexibly to divide different subspaces. Figure \ref{fig3}(a) shows the calculation of a hyperplane in three-dimensional space. The main functionality of the weights $\mathbf{w_{i}}$ is to learn the distribution of different input features instead of adjusting $H=0$. When the weights are fixed, the angle between the hyperplane $H=0$ and the coordinate system is also fixed. In this case, the subspace divided by the hyperplane is usually not optimal and the learning of weights easily converges to a local minimum. To add more representation flexibility to the hyperplane $H=0$, the standard IC neuron use adjustment weight $w'$. Then there are two independent parameters in the calculation of the hyperplane: $w'$ is used to change the angle between the hyperplane and the coordinate system, and the bias $b_{2}$ is used to translate the hyperplane in the whole space. Figure \ref{fig3}(b) shows how the angle of the hyperplane changes after adding the adjustment weight. For higher dimensions, we give a theoretical explanation of the adjustable range of $w'$:

Theorem 1. By adjusting $w'$, the hyperplane $\sum_{N}^{i=1} \ (w_{i} - w')x_{i} = 0$ can be rotated $\pi$ around the cross product of vector $\mathbf{W}$ and $\mathbf{I}$ when two vectors are linearity independent. Here $\mathbf{W} = (w_{1},w_{2},\dots,w_{N})^{T}$ and $\mathbf{I} = (\underbrace{1,1,\cdots,1}_{N})^{T}$.

\begin{proof}
	The normal vector of $\sum_{i=1}^{n} \ (w_{i} - w')x_{i} = 0$ is given by $\mathbf{H} = (w_{1}-w',w_{2}-w',\dots,w_{n}-w')^{T}$. Consider the angle between $\mathbf{H}$ and $\mathbf{I}$:
	\begin{equation}
	\label{eq7}
	\cos(\theta) = \frac{\mathbf{H}^{T} \cdot \mathbf{I}}{|\mathbf{H}| |\mathbf{I}|}
	= \frac{\mathbf{W}^{T} \cdot \mathbf{I} - Nw'}{\sqrt{N|\mathbf{W}|^{2} - 2Nw'\mathbf{W}^{T} \cdot \mathbf{I} + N^{2}w'^{2}}}. 
	\end{equation}
	When $\mathbf{W}^{T} \cdot \mathbf{I} \ge Nw'$: 
	\begin{equation}
	\label{eq8}
	\cos(\theta) = \sqrt{1 + \frac{(\mathbf{W}^{T} \cdot \mathbf{I})^{2} - N|\mathbf{W}|^{2}}{N|\mathbf{W}|^{2} - 2Nw'\mathbf{W}^{T} \cdot \mathbf{I} + N^{2}w'^{2}}}.
	\end{equation}
	Eq. \ref{eq6} is a continuous subtractive function, and $\cos(\theta) \in [0,1)$ when $w' \in (-\infty,\frac{\mathbf{W}^{T} \cdot \mathbf{I}}{n}]$. Similarly, $\cos(\theta) \in (-1,0]$ when $w' \in [\frac{\mathbf{W}^{T} \cdot \mathbf{I}}{n},\infty)$. Therefore, we get $\theta \in (0,\pi)$ when $w' \in (-\infty,\infty)$. It is clear that the direction of the rotation axis is the same as the cross product of $\mathbf{W}$ and $\mathbf{I}$.
\end{proof}

Theorem 1 implies that $\sum_{i=1}^{n} \ (w_{i} - w')x_{i} = 0$ can almost represent the whole hyperplanes parallel to the cross product of $\mathbf{W}$ and $\mathbf{I}$, providing flexible strategies for dividing spaces. In summary, by adjusting the relationship between $w$ and $w'$, the IC neuron can not only retain the representation ability of the MP neuron but also flexibly segment linear representation spaces for some complex distribution.

\subsection{Building the IC Networks}

We further build neural networks by using the standard IC neurons. Due to the characteristic that the IC neuron can transform the input with any size to 1-D output, we integrate the IC neuron into most popular neural network frameworks, such as fully-connected, convolutional, and recurrent structures. 

\textbf{Full-connected structure.} First, we consider building a full-connected layer with the IC neuron. The input size is $n$ and the number of neurons is $m$, and the output of the $j$th neuron can be expressed as:

\begin{equation}
\label{eq9}
\begin{aligned}
\centering
y_{j} &= \mathrm{f} \left( \sum_{i=1}^{n} \ w_{ij}x_{i} + \sigma \left(\sum_{i=1}^{n} \ w_{ij}x_{i} - w'_{j}x_{sum} \right) \right) , j \in [1, m] . \\
\end{aligned}
\end{equation}

To simplify the notation, bias terms are omitted here and below. The eq. 9 can be redefined by matrices and vector:
\begin{equation}
\label{eq10}
\begin{aligned}
\centering
\mathbf{y} &= \mathrm{f} \left(\mathbf{W}_{n \times m}\mathbf{x} + \sigma \left(\mathbf{W}_{n \times m}\mathbf{x} - \mathbf{w'}_{m \times 1} \mathbf{I}_{1  \times n}^{T}  \mathbf{x} \right) \right) ,\\
\end{aligned}
\end{equation}
where $I$ is defined as the transpose of a n-D all-one vector calculating the $x_{sum}$. $W$ is connection weights and $w'$ representing adjustment weights is a vector. We use eq. 10 to build the fully-connected IC network, similar to the architecture of multilayer perceptron (MLP). By hierarchically connecting the IC neurons, the IC fully-connected (IC-FC) network can represent more complex non-linear distributions. 

\textbf{Recurrent structure.}The classical RNNs use deterministic transitions from previous to current hidden states. Given an input sequence $\mathbf{x}=(\mathbf{x}_{1},\dots,\mathbf{x}_{T})$, a RNN computes the sequence of h-D hidden vectors $\mathbf{h}=(\mathbf{h}_{1},\dots,\mathbf{h}_{T})$ and sequence of output vectors $\mathbf{y}=(\mathbf{y}_{1},\dots,\mathbf{y}_{T})$ by iterating the equations:
\begin{equation}
\label{eq11}
\begin{aligned}
\mathbf{h}_{t} & = \mathrm{f} (\mathbf{W}_{n \times h}\mathbf{x}_{t} + \mathbf{W}_{h \times h}\mathbf{h}_{t-1}) , \\
\mathbf{y}_{t} & = \mathbf{W}_{h \times m}\mathbf{h}_{t} , \\ 
\end{aligned}
\end{equation}
where $\mathbf{W}$ denotes weight matrices and $\mathrm{f}$ denotes the activation function. We use the IC structure to transform the current input $\mathbf{x}_{t}$ instead of $\mathbf{W}\mathbf{x}_{t}$. When the iteration reaches the $t$th round, the calculation of $\mathbf{h}_{t}$ is defined by:
\begin{equation}
\label{eq12}
\begin{aligned}
\mathbf{h}_{t} & = & \mathrm{f} \left( \mathbf{W}_{n \times h}\mathbf{x}_{t} + \sigma(\mathbf{W}_{n \times h}\mathbf{x}_{t} -\mathbf{w'}_{h} \mathbf{I}_{n}^{T}  \mathbf{x}_{t}) + \mathbf{W}_{h \times h}\mathbf{h}_{t-1} \right) . \\
\end{aligned}
\end{equation}
We retain the linear transformation of $\mathbf{h}_{t-1}$ and the calculation of $\mathbf{y}_{t}$. Since eq. 12 retains the recurrent structure, it can learn the information of previous in sequence. Besides, the hidden state $\mathbf{h}_{t}$ emphasizing the $t$-th input $\mathbf{x}_{t}$ will make the future context easier to learn. We use eq. 12 to build the IC recurrent neural network (IC-RNN), which can process sequential data. 

\textbf{Convolutional structure.} The convolutional structure uses the kernel and sliding window to capture the feature in a local area \cite{krizhevsky2017imagenet} \cite{valueva2020application}, mapping an input feature maps $\mathbf{X} \in \mathbb{R}^{H' \times W' \times C'}$ to a output $\mathbf{U} \in \mathbb{R}^{H \times W \times C}$. To simplify the notation, the activation operator is omitted. We can write the output feature maps as $\mathbf{U}=[\mathbf{U_{1}}, \mathbf{U_{2}},\dots,\mathbf{U_{C}}]$, where  
\begin{equation}
\label{eq13}
\begin{aligned}
\mathbf{U}_{i} = \mathbf{W}_{i}*\mathbf{X}. \\
\end{aligned}
\end{equation}
Here $\mathbf{W}_{i} \in \mathbb{R}^{k \times k \times C'}$ is the set of filter kernels and $*$ is used to denote the convolutional operator. We retain the characteristics of sliding window which takes advantages of shared weight, then using eq. 5 to changing the calculation of the kernel in each window. the new kernel is represented by $[\mathbf{W_{i}}, w'_{i}]$:
\begin{equation}
\label{eq14}
\begin{aligned}
\mathbf{U_{i}} &= [\mathbf{W}_{i}, w'_{i}]*\mathbf{X}\\
&= \mathbf{W}_{i}*\mathbf{X} + \sigma(\mathbf{W}_{i}*\mathbf{X} - w'_{i} \times (\mathbf{I}*\mathbf{X})), \\ 
\end{aligned}
\end{equation}
where $\mathbf{I}$ is an all-one tensor with the same size as $\mathbf{W}_{i}$ used to sum the local area. Since eq. 14 termed the IC convolutional layer takes advantage of the dividing hyperplane, and it can capture more abstract local features without changing the format of input and output feature maps. We stack this structure to build the IC convolutional neural network (IC-CNN). Note that the pooing operation is not changed in our IC-CNNs.

\subsection{Parameters and Complexity Analysis}

\textbf{Parameters.} We compare the standard IC neuron with MP neuron in three network structures. For the IC fully-connected and recurrent structures, the addition of parameters depends completely on the adjustment weight $w'$ and bias $b_{1}$. If there are $m$ hidden neurons and n-D input, the IC networks will add parameters from $(n+1) \times m$ to $(n+3) \times m$. The ratio of addition is $\frac{2}{n}$, which can be regarded as adding $2$ dimensions to the input. The result of the convolutional structure is similar in that the addition depends on the number of kernels. The real-world tasks usually need hundreds of neurons or kernels, making the effect of additional parameters of the IC neuron negligible.

\textbf{Computational complexity.} The computational complexity of neural networks mainly comes from matrix operations. In the IC fully-connected and recurrent structures, $\mathbf{W}\mathbf{x}$ term calculating $(n \times m) \times (m \times 1)$ is only calculated once. $\mathbf{w'} \mathbf{I}^{T}  \mathbf{x} = \mathbf{w'} (\mathbf{I}^{T}  \mathbf{x})$ needs to calculate $(1 \times n) \times (n \times 1)$ and $(M \times 1) \times (1 \times 1)$. Compared to the hidden layer with MP neurons. The additional complexity is approximately $\frac{1}{n}+\frac{1}{m}$. As for convolutional structure, $\mathbf{W}_{i}*\mathbf{X}$ is also calculated once. Note that we can first generate a tensor to represent $w'_{i} \times \mathbf{I}$. $w'_{i} \times (\mathbf{I}*\mathbf{X}) = (w'_{i} \times \mathbf{I})*\mathbf{X}$ can be regarded as adding a kernel for input features. The additional complexity is approximately $\frac{1}{C}$. We do not change the operations outside the neuron model, include batch normalization, pooling, etc. When $N$, $N$ or $C$ are relatively large, the complexity of IC networks is similar to the networks with MP neurons.

\section{Experiment}
This section evaluate the performance of the IC networks by integrating the IC neurons into three network structures (IC-FC, IC-rnn and IC-CNN). 

\subsection{Configurations and Datasets}

Our goal is to compare the IC networks with traditional ones. For fair comparison, we evaluate the traditional networks with the same network hyperparameters as IC networks. Specifically, we use the ReLU as the activation function $\mathrm{f}$, cross entropy for loss function and adadelta for optimizer. The evaluation metrics include top-1 accuracy, FLOPs and parameter amount. The accuracy of each round is recorded to trace the training process. During the training process, the test accuracy fluctuates sharply. We make a simple smoothing of the test curve in order to better show the trend of test accuracy changes. Besides, we evaluate the basic version of the IC neuron (eq~\eqref{eq4}) to investigate the effectiveness of the $\sigma$ operation and the adjustment weight $w'$. These results are represented by $-B$ version in our tables. We repeat each experiment three times and record the best accuracy in order to reduce the influence of random initialization. 

We use a variety of classification tasks to evaluate the different models. For each tasks, we individually set the number of layers and hidden neurons. The details are shown below.


\textbf{Sentiment classification} task uses the IMDB dataset that contains $25000$ movie reviews for training and $25000$ for testing. The reviews are preprocessed by extracting $tf-idf$ features and are cropped into $5000$-dimensional features of equal length. We build the IC-FC network with the input-1024-512-256-output structure. 

\textbf{Music classification} task uses the GTZAN dataset that contains $1000$ music clips divided into $10$ classes evenly. Each clip is stored in a $30$ seconds long file. We split the dataset into $710$ clips for training and $290$ clips for testing. In addition, we use MFCC, chroma, mel-spectrogram, spectral contrast and tonnetz feature to represent each music clip, which transform the original sound wave into a $253$ feature vector. We build the IC-FC network with the input-256-128-output structure. 

\begin{table*}
	\caption{The accuracy of three FC models on several classification tasks.}\label{tbl1}
	\setlength{\tabcolsep}{6mm}{
	\begin{tabular}{lccccccc}
		\toprule
		Model   & MNIST & IMDB & GTZAN & sEMG & LETTER & YEAST & ADULT \\
		\midrule
		FC      & 98.02  & 87.59 & 54.14 & 40.00 & 82.53 & 57.62 & 84.99 \\
		IC-FC   & 98.42  & 88.28 & 56.21 & 41.67 & 85.70 & 61.04 & 85.20 \\
		IC-FC-B & 98.37  & 88.13 & 55.17 & 41.11 & 84.13 & 60.10 & 85.14 \\
		\bottomrule
	\end{tabular}}
\end{table*}

\begin{table*}
	\caption{The FLOPs/storages of three FC models on several classification tasks. The FLOPs (KMac) denote the amount of calculation for feedforward propagation. The storages (KB) reflect the amount of learnable parameters.}\label{tbl2}
	\setlength{\tabcolsep}{2mm}{
	\begin{tabular}{lccccccc}
		\toprule
		Model   & MNIST & IMDB & GTZAN & sEMG & LETTER & YEAST & ADULT \\
		\midrule
		FC      & 235.14/235.15  & 5777.66/5777.67 & 99.20/99.21 & 3600.90/3600.92 & 4.00/4.03 & 0.98/0.99 & 1.04/1.04 \\
		IC-FC   & 236.56/235.91  & 5785.99/5781.25 & 100.09/99.98 & 3606.46/3603.97 & 4.18/4.22 & 1.06/1.08 & 1.13/1.14 \\
		IC-FC-B & 236.18/235.53  & 5784.20/5779.46 & 99.71/99.59 & 3604.92/3602.44 & 4.08/4.12 & 1.02/1.03 & 1.09/1.09 \\
		\bottomrule
	\end{tabular}}
\end{table*}

\textbf{Hand movement recognition} task uses the uci sEMG dataset that contains $1800$ records divided into six kinds of hand movements, i.e., spherical, tip, palmar, lateral, cylindrical and hook. The sEMG dataset is a time-series dataset, where EMG sensors capture $500$ features per second and each record associated with $3000$ features. We build the IC-FC network with input-1024-512-output to process the records. We use the IC-RNN with an $input-80-output$ structure to recurrently process the each EMG features of records. 

\textbf{Speech recognition} task uses the Google speech commands dataset (V1) \cite{warden2018speech} that contains $65000$ one-second long utterances of $30$ short words. The researchers often use $20$ common commands dataset to simplify the task, which is named 20-cmd. We use 20-cmd to further evaluate the IC-RNN with an $input-64-output$ structure.

\textbf{Image classification} task uses the MNIST and CIFAR-10 dataset. The MNIST contains $60K$ training images of $28 \times 28$ size and $10K$ test images. We evaluate the IC-FC network with an input-256-128-output structure. The IC-CNN for MNIST is consist of three $3 \times 3$ IC convolutional layer with $32,64,64$ output channels and $1,2,2$ strides, respectively. After the IC convolutional layers capture the features, we use a features-256-output transitional FC structure as classifier. We use CIFAR-10 to further evaluate this IC-CNN. The CIFAR-10 is consists of $60K$ $32 \times 32 \times 3$ RGB images in $10$ classes divided into $50K$ training images and $10K$ testing images.

Besides, we evaluate some simple IC networks on the uci YEAST, ADULT and LETTER datasets, which have only $8/14/16$ dimensional inputs. YEAST has $1038/446$ training/test examples. LETTER has $16000/4000$ training/test examples. ADULT has $32561/16281$ training/test examples. We build the IC-FC networks with the input-32-16-output, input-32-16-output, input-64-32-output for YEAST, ADULT and LETTER datasets, respectively. Note that the variance of ADULT data is relatively big, so we set a BN operation at the beginning of the networks. 

\subsection{Comparative Experiment}

\subsubsection{IC-FC Networks}

\begin{figure*}[]
	\centering
	\subfigure[MNIST]{
		\includegraphics[scale=0.4]{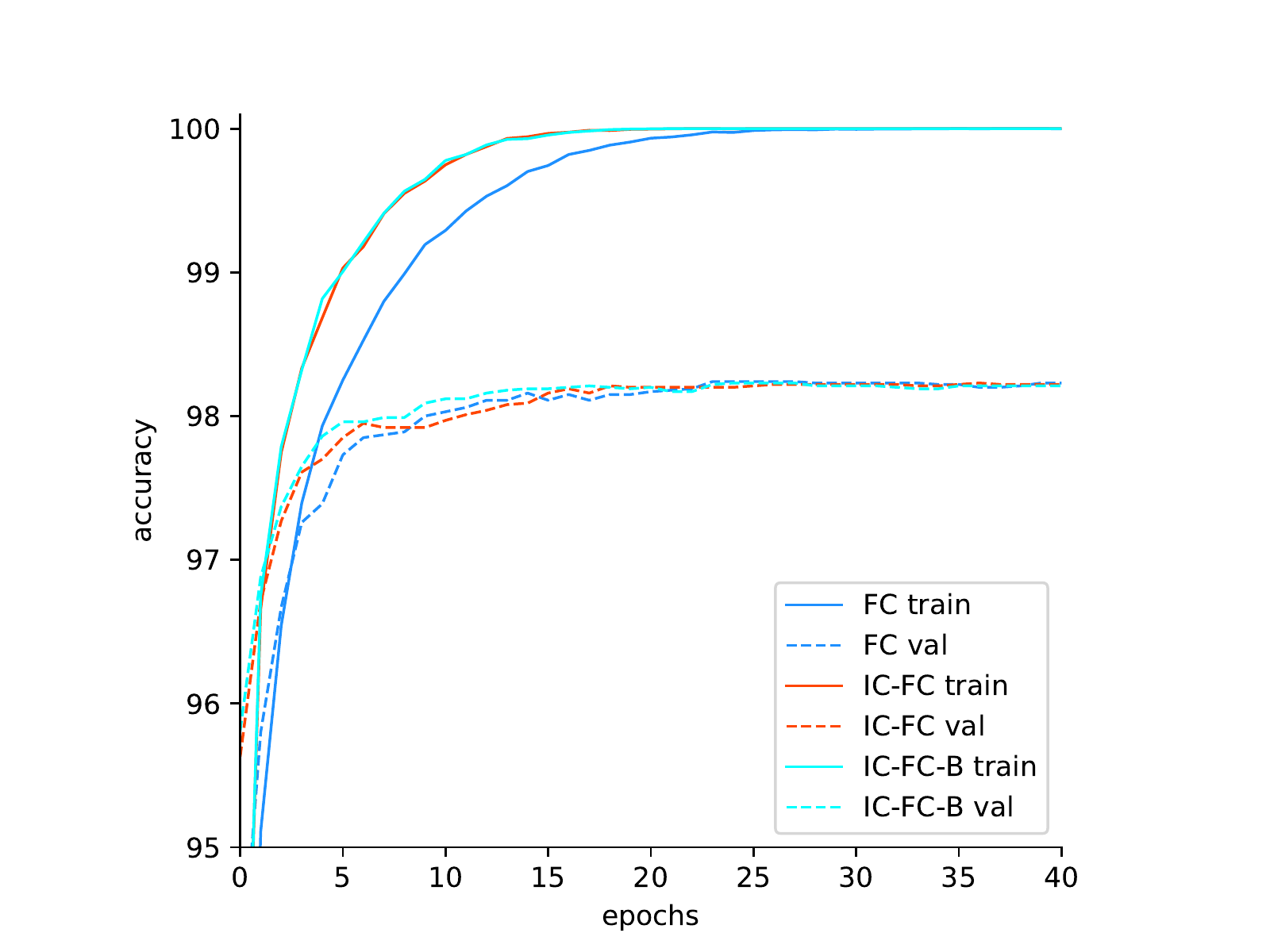}
	}
	\subfigure[IMBD]{
		\includegraphics[scale=0.4]{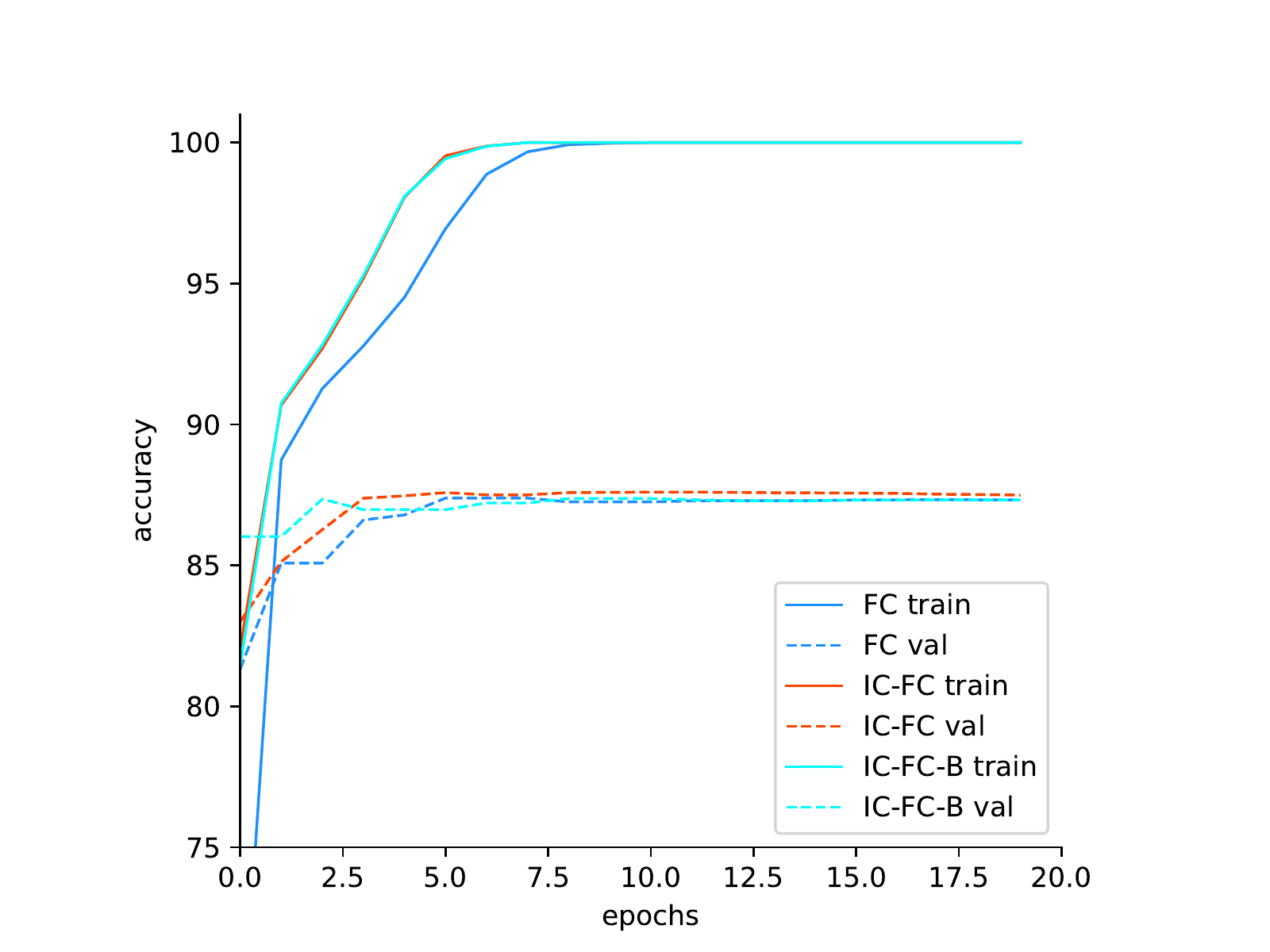}
	}
	\subfigure[GTZAN]{
		\includegraphics[scale=0.4]{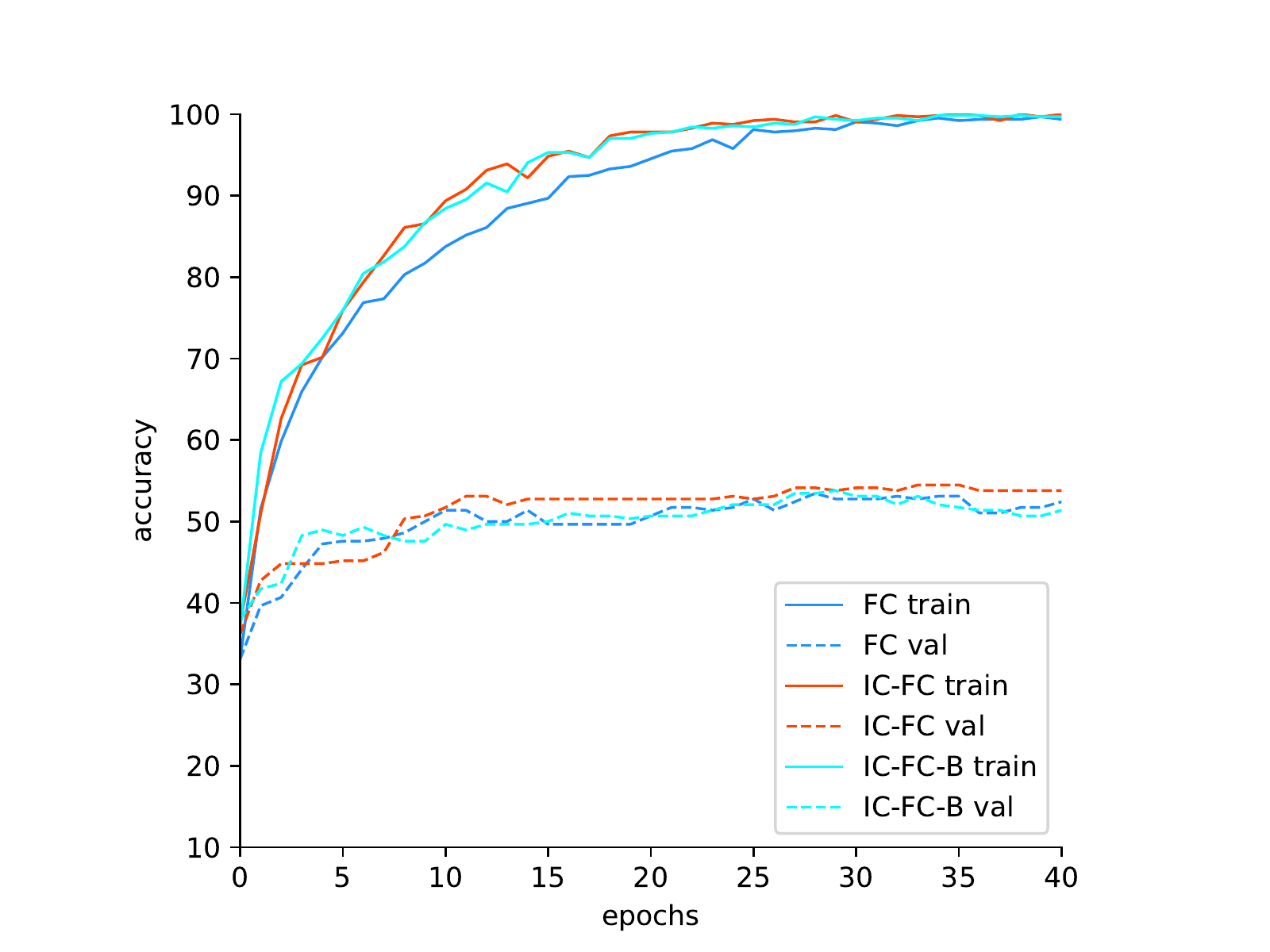}
	}
	\subfigure[sEMG]{
		\includegraphics[scale=0.4]{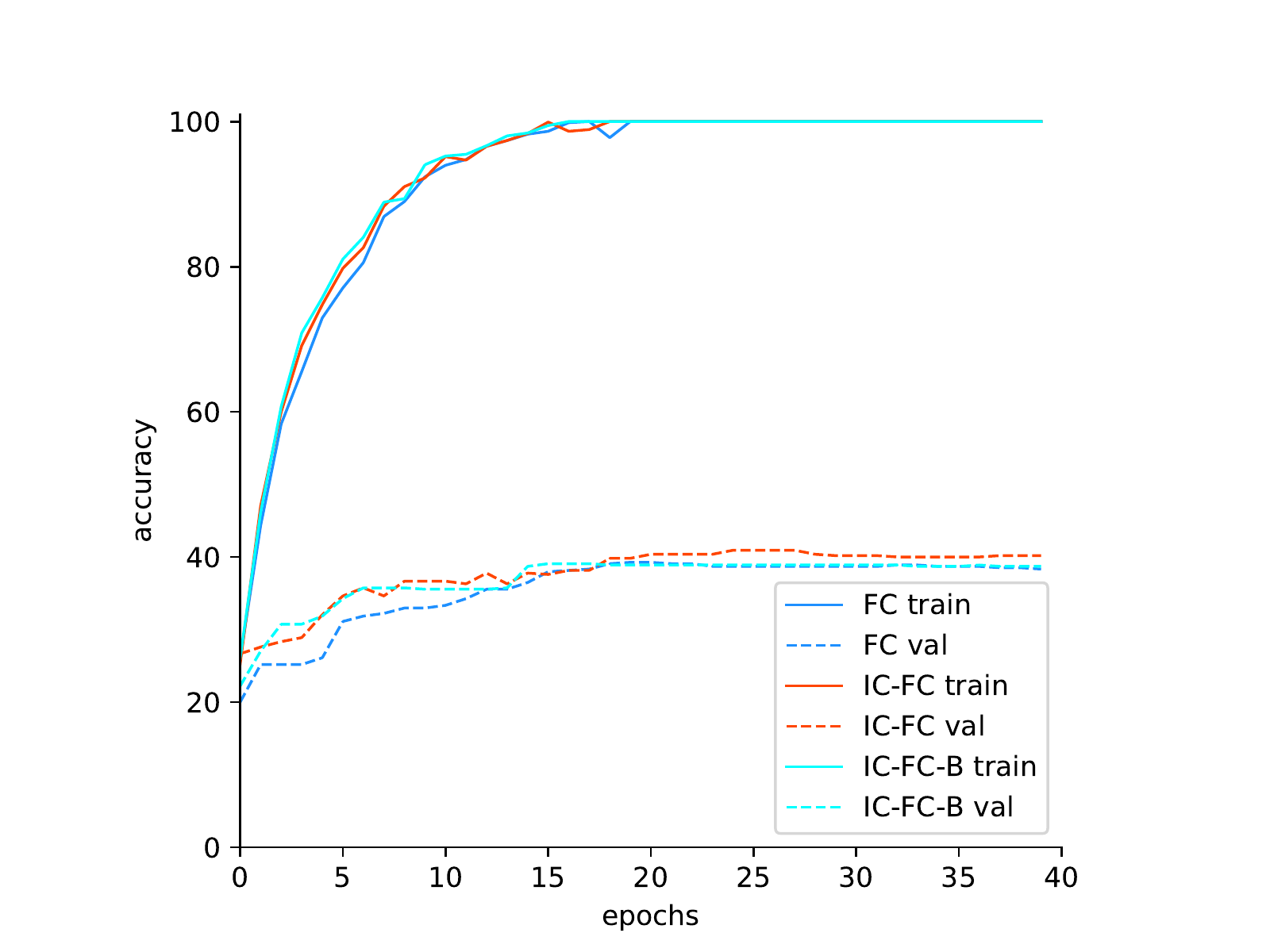}
	}
	\subfigure[LETTER]{
		\includegraphics[scale=0.4]{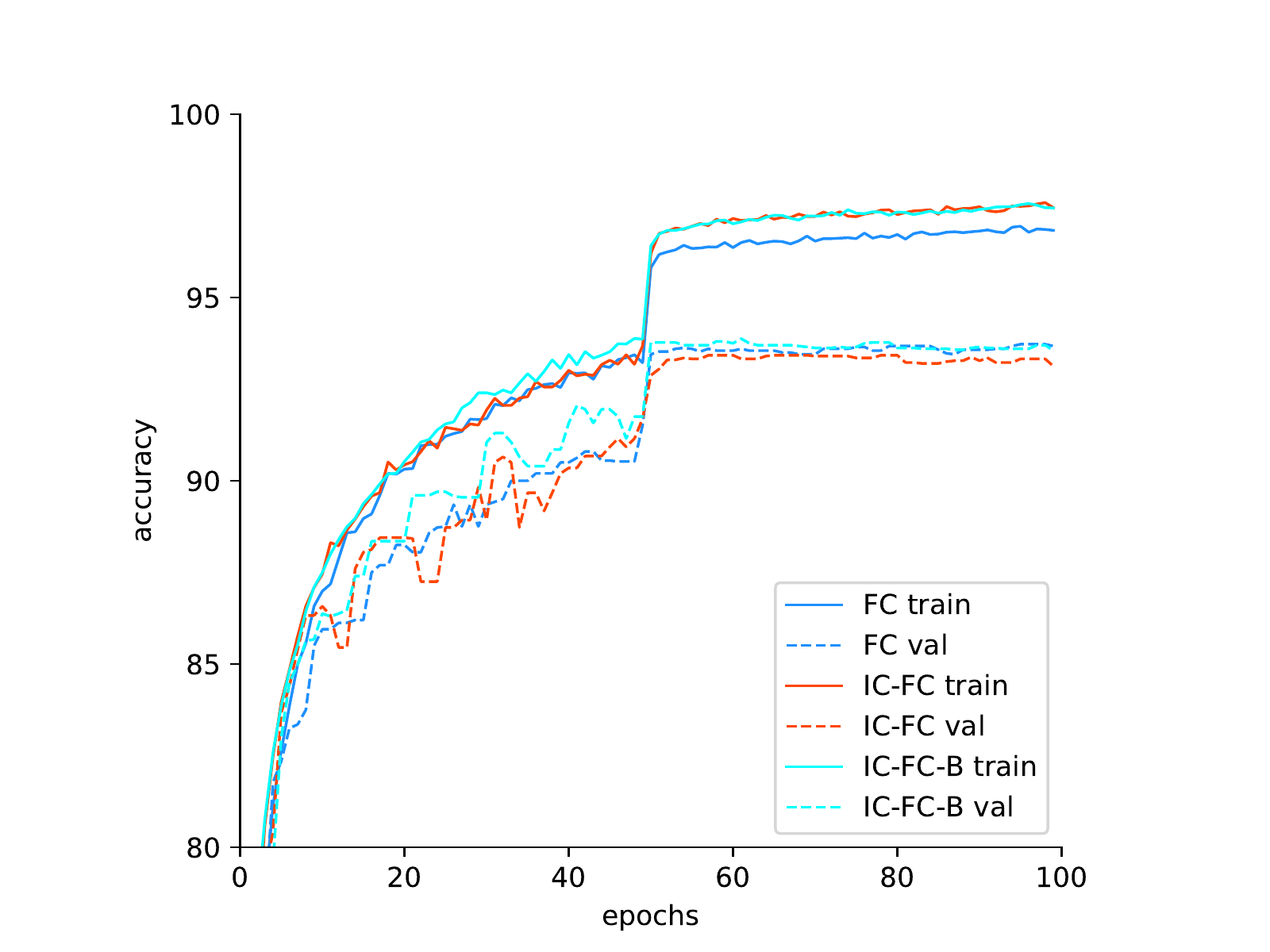}
	}
	\subfigure[YEAST]{
		\includegraphics[scale=0.4]{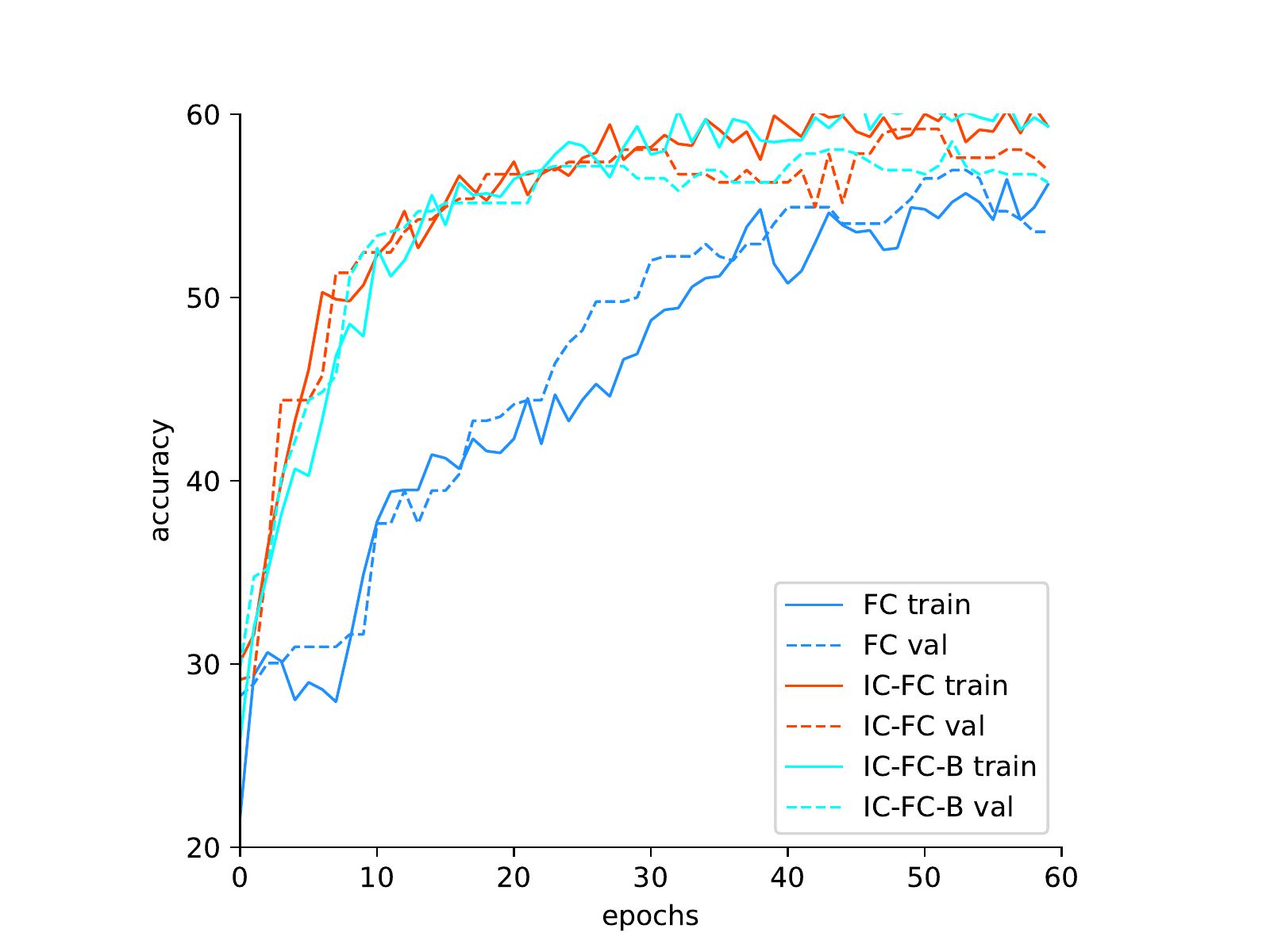}
	}
	
	\subfigure[ADULT]{
		\includegraphics[scale=0.4]{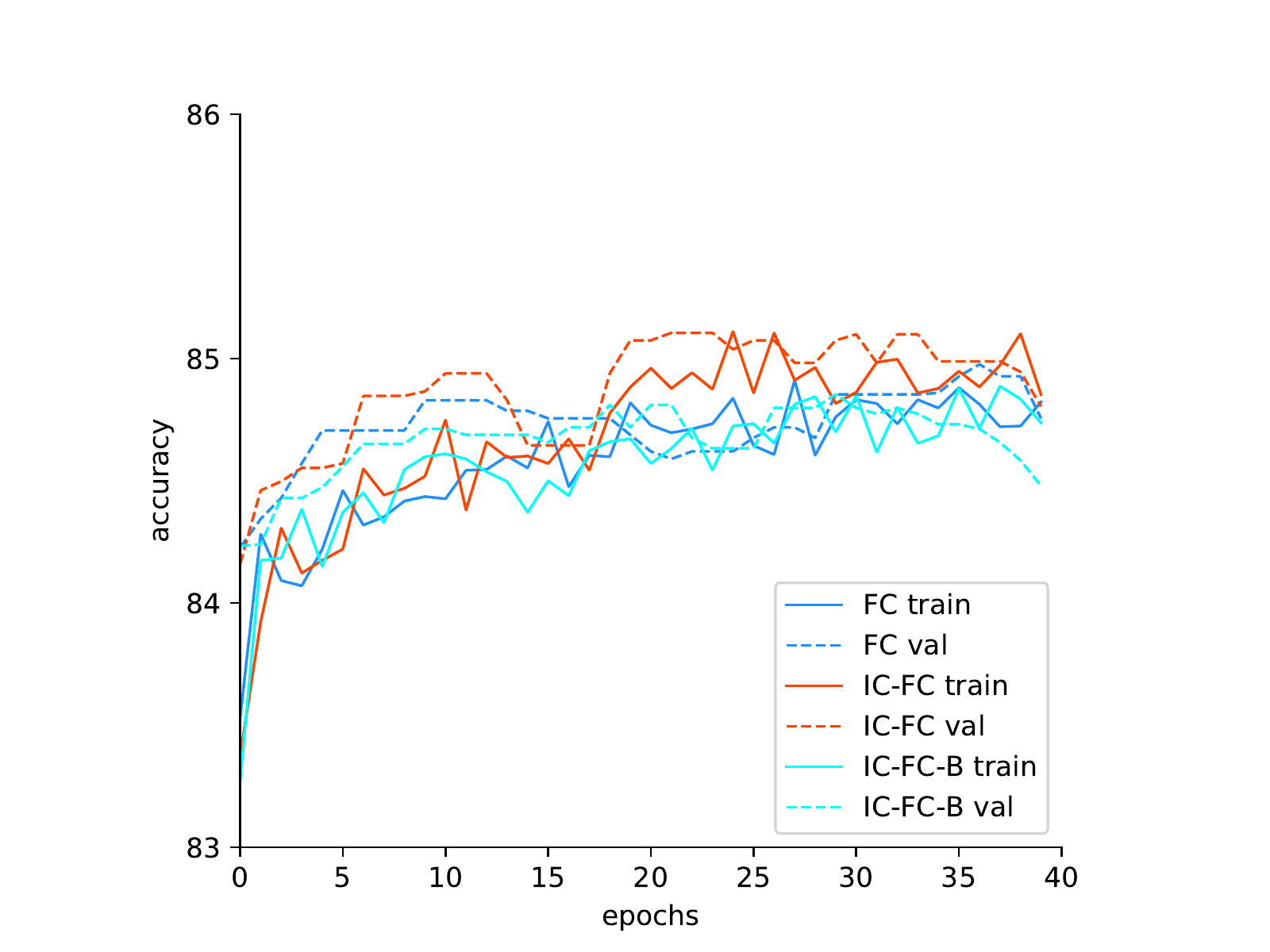}
	}
	\caption{The training curves of the IC-FC networks and the FC networks on several task. }
	\label{fig4}
\end{figure*}

Table 1 records the performance of IC-FC networks and corresponding FC networks on the above tasks except the CIFAR-10 and Google speech commands. We have three major observations. First, the IC networks achieve highly competitive performance in all the tasks. Remarkablely, the IC networks have a obvious improvement when the accuracy of original networks is low, such as GTZAN ($2.07\%$), sEMG ($1.67\%$) and YEAST ($3.37\%$). More importantly, we observe that the IC-FC networks add little computational cost from Table 2. The most obvious addition of FLOPs and storages is the results on the YEAST ($8.16\%$/$9.09\%$) and ADULT ($8.65\%$/$9.62\%$) dataset. According to our analysis in section 2.5, the influence of calculation and parameters will decrease as the number of hidden-layer neurons increases. This conclusion is proven by the IC networks on IMDB and sEMG datasets, which bring negligible additional cost compare to FC networks ($0.14\%$/$0.06\%$ and $0.15\%$/$0.08\%$). In summary, the IC-FC networks show the better representation ability with little additional computational burden. Besides, we observe that the standard version of the IC neuron has an improvement compared to the basic one in most tasks. This result shows that the adjustment weight can work well through joint training with the other learnable parameters. 

The training processes of several FC models are depicted in Fig. 4. It is clear that the IC networks converge in a short time, indicating that IC neurons are more conducive to fitting features. Specially, although both the IC networks and FC networks use the same initialization and optimizer, the IC networks fit the distributions of training datasets better in the first few rounds. When the training accuracy is overly high, the improvement of the test curves is not obvious in the GTZAN and sEMG datasets. We argue that this is because of overfitting. Although the IC neuron help to find the fine-gain features, the type of these features is similar to ones got from the MP neuron. Besides, two versions of the IC neuron show similar convergence speeds, showing that the strong representation ability comes from the $\sigma$ non-linear operation of the IC neuron. Moreover, the adjustment weight $w'$ can improve the upper bound of representation ability.

\subsubsection{IC-RNN and IC-CNN}

We build the IC-RNNs to process time-series data, including sENG and Google speech commands datasets (20-cmd). Since the dropout does not work well with RNNs and large RNNs tend to overfit~\cite{zaremba2014recurrent}, we build the IC-RNNs and traditional RNNs with the small scale. The results are recorded in Table 3, showing that the IC-RNNs achieve the much higher accuracy. The IC-RNNs use a basic recurrent structure, which shows that the IC neuron can combine with the recurrent structure well. Note that the accuracy of Google speech commands datasets is lower than original work~\cite{de2018neural}, since we use the simplest recurrent models. We believe that the IC neuron can improve the complex recurrent models, including LSTM~\cite{sak2014long}, etc. 

\begin{figure*}[]
	\centering
	\subfigure[sEMG]{
		\includegraphics[scale=0.4]{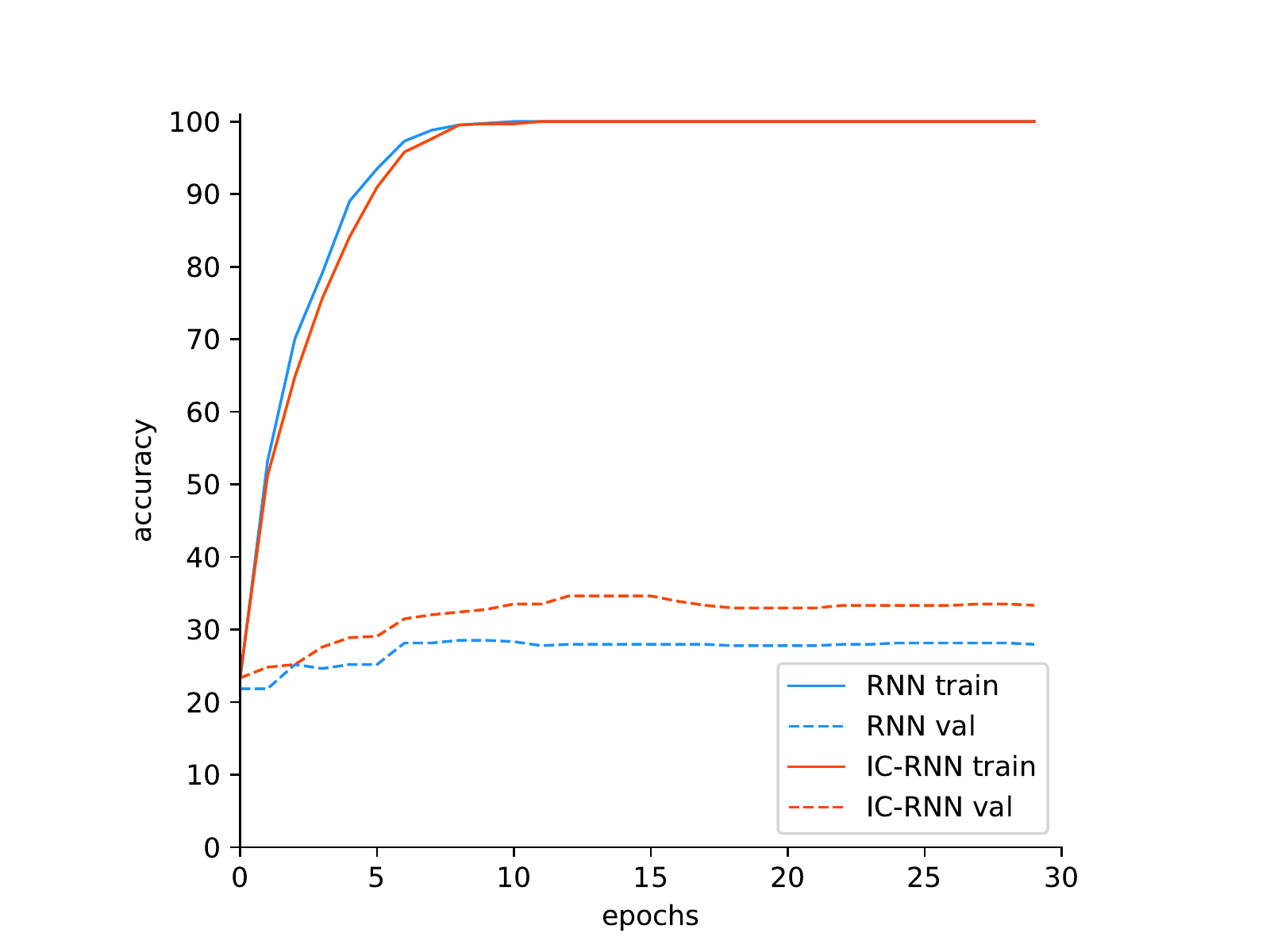}
	}
	\subfigure[20-cmd]{
		\includegraphics[scale=0.4]{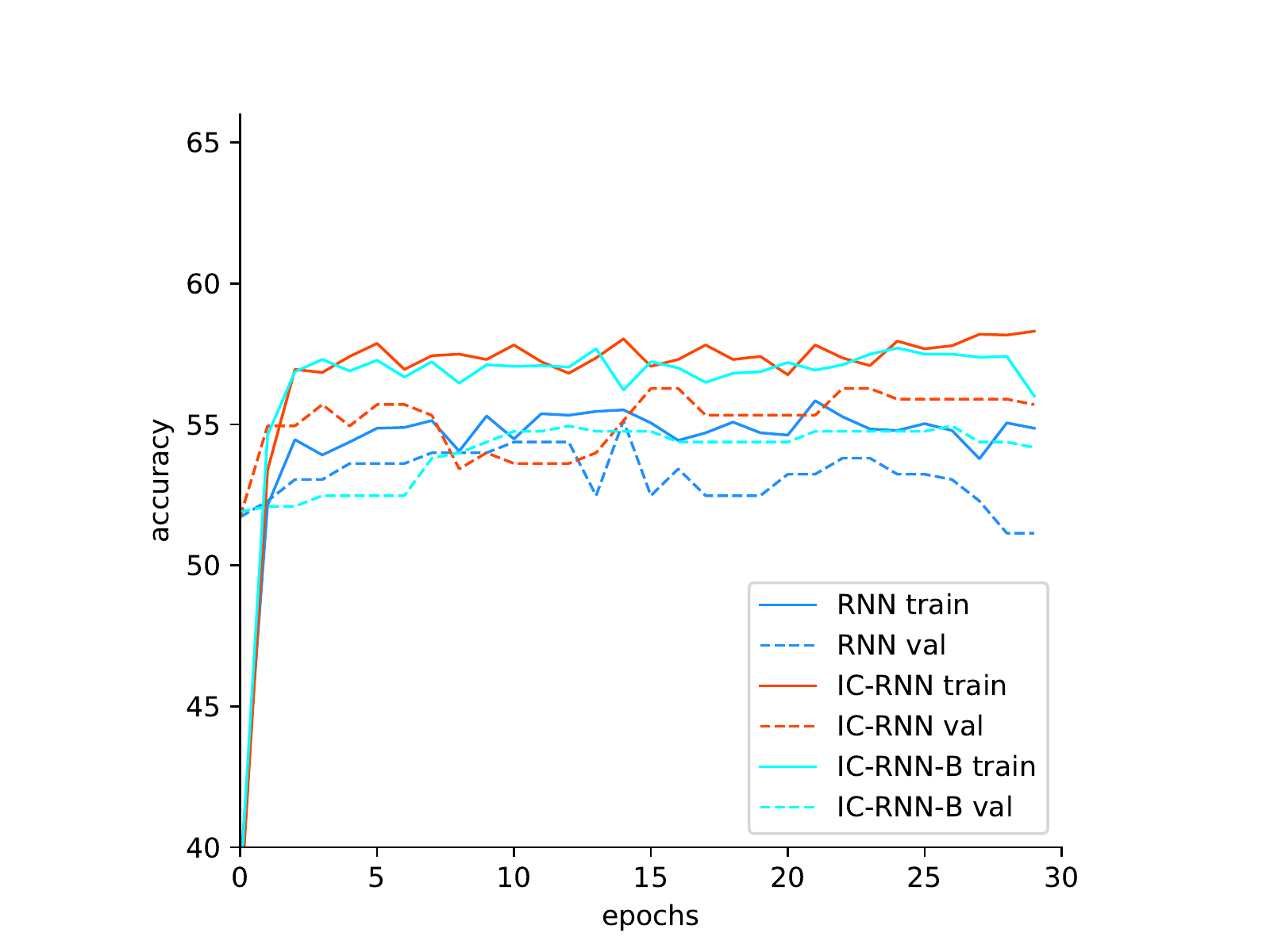}
	}
	\caption{The training curves of the IC-RNNs and the RNNs on tmie-series tasks. }
	\label{fig5}
\end{figure*}

\begin{figure*}[]
	\centering
		\subfigure[MNIST]{
			\includegraphics[scale=0.4]{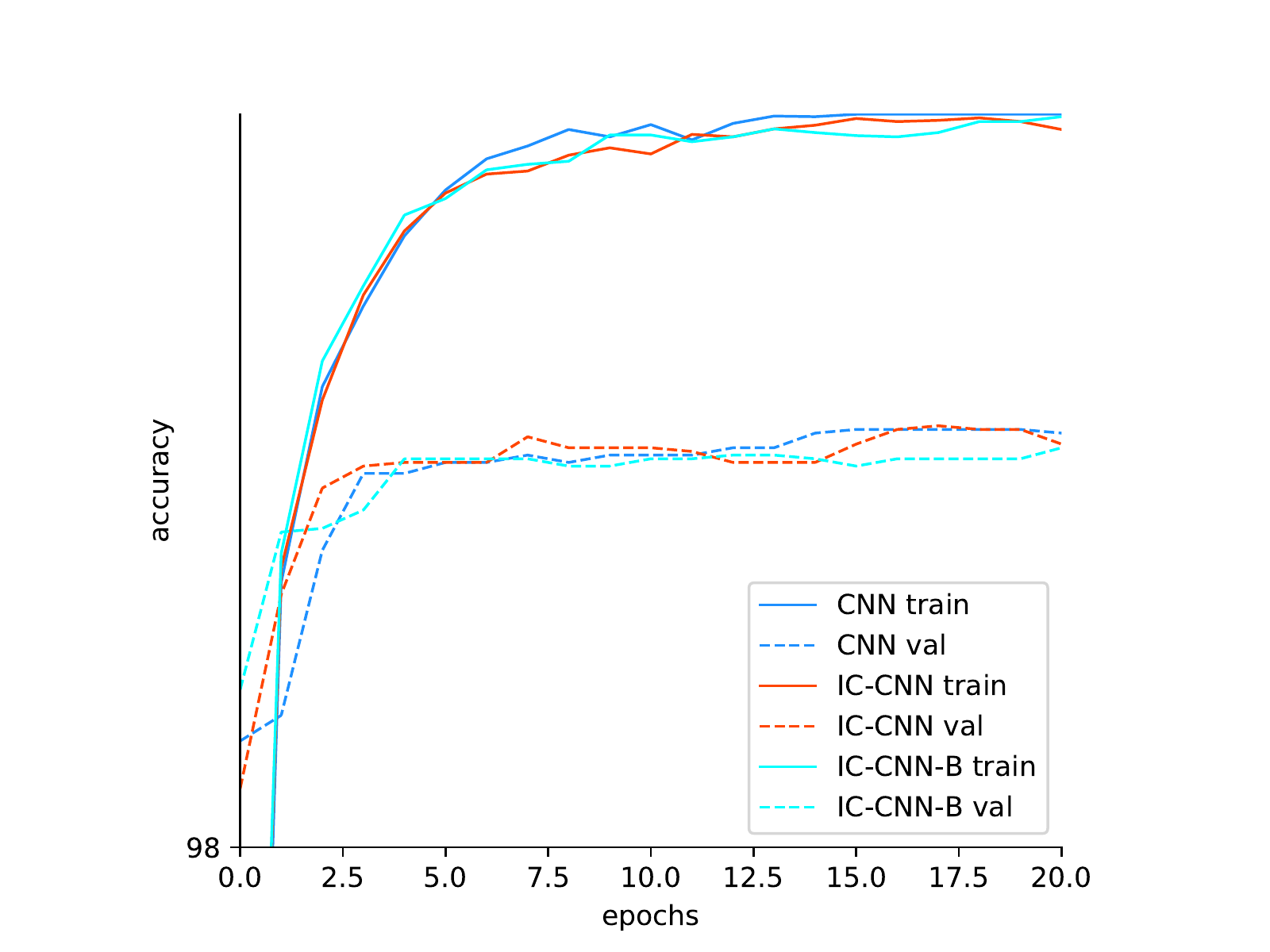}
		}
	\subfigure[CIFAR-10]{
		\includegraphics[scale=0.4]{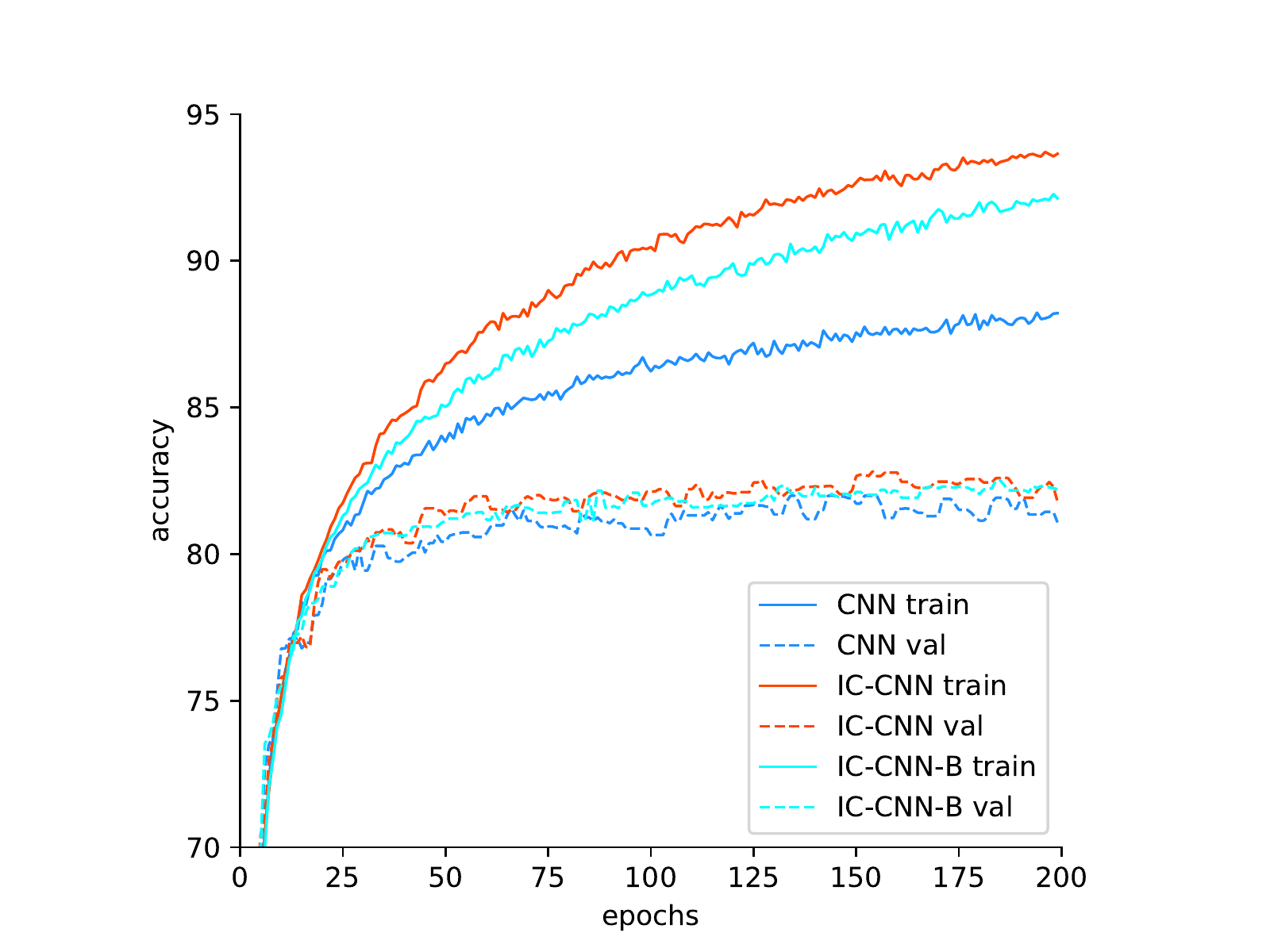}
	}
	\caption{The training curves of the IC-CNNs and the CNNs on image tasks. }
	\label{fig6}
\end{figure*}

\begin{table}
	\caption{The accuracy/FLOPs/storages of the IC-RNNs and RNNs on tmie-series tasks..}\label{tbl3}
	\begin{tabular}{lcc}
		\toprule
		Model   & sEMG & 20-cmd \\
		\midrule
		FC      & 28.95/240.48/240.57  & 57.98/662.46/662.56 \\
		IC-FC   & 31.42/243.64/240.73  & 59.13/672.91/662.69  \\
		IC-FC-B & 30.86/243.56/240.65  & 59.10/672.85/662.62  \\
		\bottomrule
	\end{tabular}
\end{table}

\begin{table}
	\caption{The accuracy/FLOPs/storages of the IC-CNNs and CNNs on image tasks.}\label{tbl4}
	\begin{tabular}{lcc}
		\toprule
		Model   & MNIST & CIFAR-10 \\
		\midrule
		FC      & 98.84/6491.20/861.39 & 81.25/9067.26/1107.72 \\
		IC-FC   & 99.12/6623.70/861.71 & 82.49/9258.75/1108.05 \\
		IC-FC-B & 99.02/6582.93/861.55 & 82.82/9205.50/1107.88 \\
		\bottomrule
	\end{tabular}
\end{table}

We build the IC-CNN to process image data, including the MNIST and CIFAR-10 datasets. To only evaluate the IC neuron in convolutional stucture, the IC-CNNs use the traditional full-connected layers as classifiers instead of the IC-FC. It is well know that kernels can capture the local feature through convolution operation. The results of the IC-CNNs recorded in Table 4 achieve a higher accuracy, implying that the IC neurons learning the more fine-gain local features. The training curves shown in Fig.6 show a faster convergence speed of the IC-CNNs. These comparison results show that the IC neuron work well on basic convolutional structure. Similar to the conclusion of the IC-RNNs, we believe that the IC neuron is useful unit to build bigger convolutional networks.

In summary, our experiments show that the combination of the IC neuron and three computational structures is successful. The IC networks can capture the fine-gain feature, making themselves have a higher accuracy and a faster convergence speed. However, they only increase an small computational cost compare the classical networks. We think that using the IC neuron is more efficient than expanding the scale of networks. To prove our conjecture, we compare the IC networks and bigger classical networks. Besides, we further investigate the factors affecting training process. This series of experiments is recorded in section 3.3. 

\subsection{Model capacity and performance}

We construct the experiments to prove the conjecture that using the IC neuron is better than simply increasing the parameters. To consider both high-dimensional and low-dimensional data, we evaluate the IC-FC networks on the IMDB and the YEAST datasets. The bigger full-connected networks are: input-1500-1000-500-input and input-1024-1024-512-256-output for IMDB; input-32-32-16-output and input-100-50-output for YEAST. We named them deeper or wider FC networks. The results and training curve are recorded in Table 5. We observe that expanding the scale of networks is inefficient. Although the deeper and wider FC networks use much more parameters and FLOPs, thier results are still under the IC networks. Remarkablely, the wider FC network for YEAST cost six times the computational source than the IC-FC network (6.45KMacs/6.46KB vs. 1.06KMacs/1.08KB) while its accuracy is still under the IC-FC network (60.76 vs. 61.04). Besides, some work stated that large-scale networks training is slow~\cite{ioffe2015batch}. We also observe that expanding the scale has no effect on training speed. The IC networks have strong advantage in this respect.

\subsection{Activation function}

Another supplementary experiment is to investigate the influence of different activation operation. We have mentioned that the $\mathrm{f}$ of the IC neuron can be most activation functions. However, the neuron model may be hard to train when the non-linear form is complex. We second investigate the effect of the common used function, including Tanh~\cite{nwankpa2018activation} and ELU~\cite{clevert2015fast}. They can represent smooth boundaries and negative values. From Table 6, we observe that the IC networks have an improvement with other activation functions. These experiments show that the IC neuron is flexible and universal to replace the MP neuron in many commonly used networks.

\subsection{Discussion}

In summary, a network composed of IC neurons can achieve a generalization performance better than that of MP neurons in a wide range of tasks. Experiments show that this improvement does not come from the increase in the number of calculations and parameters. On the one hand, compared to traditional networks that use the same hyperparameters, the additional computational burden brought by the IC network is basically negligible. On the other hand, we find it difficult for deeper and wider traditional networks to surpass the accuracy of IC networks. We think this is because IC neurons have more non-linear representation capabilities and can fit more fine-grain features. In the training process, this advantage also speeds up the convergence speed. Besides, the combination of multiple classification tasks and three network structures also confirmed that IC neurons are a universal computing unit. We believe that IC neurons can be combined with many existing networks to achieve good performance.

\begin{table}
	\caption{The comparison betwwenn the IC-FC networks and bigger FC networks. The value of each grid represents accuracy/FLOPs/storages.}\label{tbl5}
	\begin{tabular}{lcc}
		\toprule
		Model   & IMDB  & YEAST   \\
		\midrule
		Deeper FC& 87.98/6811.89/6811.91 & 58.74/2.03/2.04 \\
		Wider FC & 87.74/9504.01/9504.04 & 60.76/6.45/6.46 \\
		IC-FC    & 88.28/5785.99/5781.25 & 61.04/1.06/1.08  \\
		\bottomrule
	\end{tabular}
\end{table}


\section{Conclusion}

Inspired by the elastic collision model, we propose the IC neuron model that could be used to fit more complex distributions compared to the MP neuron. We build the IC networks by integrating the IC neurons into the popular network structures, including the FC, convolutional and recurrent structures. Experiments a variety of classification tasks show that the IC networks reaches a higher accuracy and a faster convergence speed compared to the classical networks. Besides, the IC networks bring little additional computational burden, showing the superiority of the IC neuron. Our work provides a basic unit to build effective neural networks.

In this work, we have not fully explored the applicability to other networks that the IC neuron potentially enables. Our future work includes applications to other architectures.

	\bibliographystyle{named}
	\bibliography{arxiv_icn}
	
\end{document}